\documentclass[10pt,twocolumn,letterpaper]{article}

\usepackage[pagenumbers]{wacv} 
\usepackage{tabularx}

\definecolor{wacvblue}{rgb}{0.21,0.49,0.74}
\usepackage[pagebackref,breaklinks,colorlinks,allcolors=wacvblue]{hyperref}

\usepackage{multirow}

\usepackage[normalem]{ulem}
\usepackage[table]{xcolor}
\usepackage{booktabs}
\usepackage[T1]{fontenc}

\def\wacvPaperID{2100} 
\def\confName{WACV}
\def\confYear{2026}

\title{AusSmoke meets MultiNatSmoke: \\
a fully-labelled diverse smoke segmentation dataset}

\author{
Weihao Li \quad Hongjin Zhao \quad Gao Zhu \quad Ge\mbox{-}Peng Ji   \quad Nicholas Wilson \\[0.4ex]
Marta Yebra \quad Nick Barnes \\ [1.0ex]
Bushfire Research Centre of Excellence, Australian National University
}

\begin{document}

\twocolumn[{%
    \renewcommand\twocolumn[1][]{#1}
    \maketitle
    \begin{center}
        \includegraphics[width=0.9\textwidth,height=0.43\textheight]{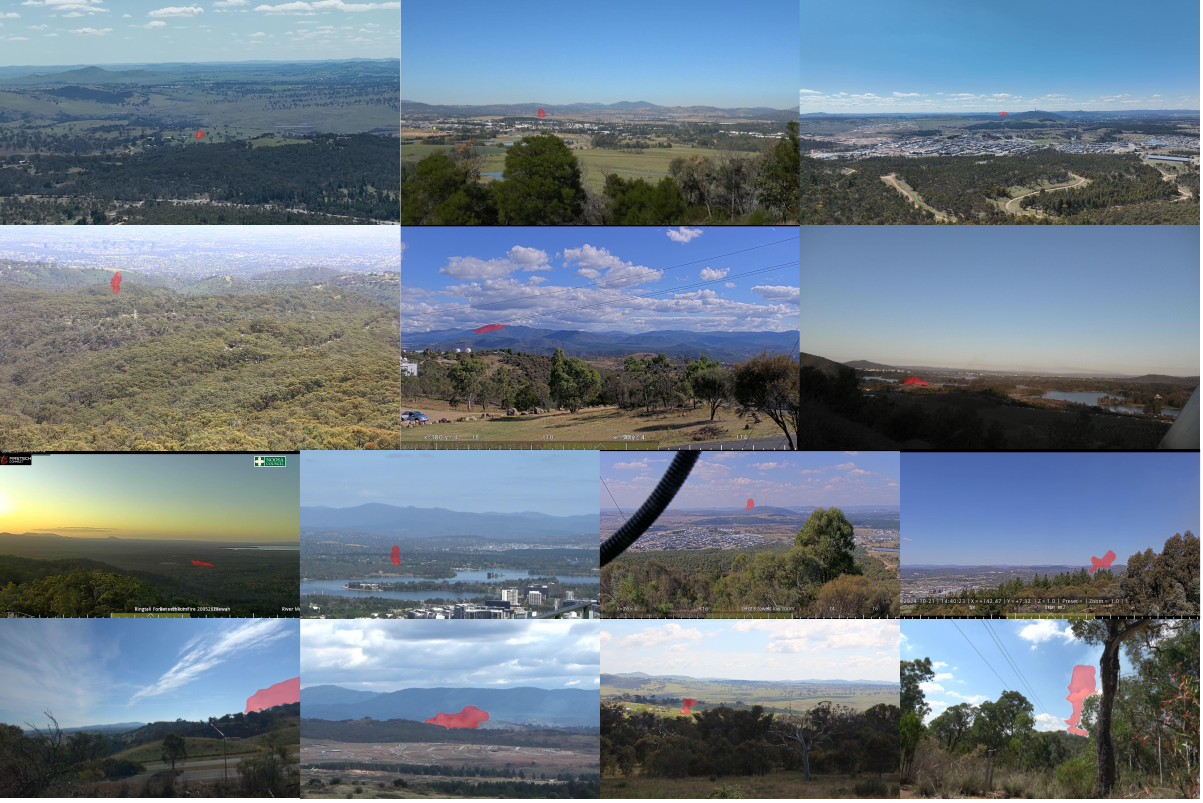}
        \captionof{figure}{Example images on wildfire smoke segmentation from our AusSmoke dataset.}
        \label{fig:teaser}
    \end{center}
}]

\begin{abstract}

Wildfires are an escalating global concern due to the devastating impacts on the environment, economy, and human health, with notable incidents such as the 2019-2020 Australian bushfires and the 2025 California wildfires underscoring the severity of these events. AI-enabled camera-based smoke detection has emerged as a promising approach for the rapid detection of wildfires. However, existing wildfire smoke segmentation datasets that are used for training detection and segmentation models are limited in scale, geographically constrained, and often rely on synthetic imagery, which hinders effective training and generalization. To overcome these limitations, we present AusSmoke, a new smoke segmentation dataset collected from Australia to address the data scarcity in this region. Furthermore, we introduce a MultiNational geographically diverse and substantially larger fully-labelled benchmark, called MultiNatSmoke, that consolidates publicly available international datasets with the newly collected Australian imagery, expanding the scale by an order of magnitude over previous collections. Finally, we benchmark smoke segmentation models, demonstrating improved performance and enhanced generalization across diverse geographical contexts.
The project is available at \href{https://github.com/henryzhao0615/MultiNatSmoke}{Github}.
\end{abstract}
    
\newpage
\section{Introduction}

Wildfires are increasingly becoming a major global concern due to their severe environmental, health, and economic consequences. Two of the most catastrophic events in recent memory include the 2019–2020 Black Summer bushfires in Australia and the 2025 wildfires in Los Angeles. The Black Summer fires burned a record 19 million hectares, destroyed more than 3,000 homes, displaced tens of thousands of people, and are estimated to have killed billions of animals \cite{boer2020unprecedented,NHRA2023}. The 2025 Los Angeles wildfires destroyed over 10,000 homes, claimed the lives of more than two dozen people, and highlighted the urgent need for holistic fire management strategies \cite{Fitch2025}.  Camera-based early smoke detection has the potential to significantly improve response times by identifying ignitions when they are still small and more easily contained, reducing overall fire impact \cite{yebra2024technological}.

Novel datasets \cite{deng2009imagenet,lin2014microsoft,li2018deep,ji2026frontiers,ji2022video,Zhao2026DermEVAL,ji2025colonx} play a crucial role in driving progress and innovation in artificial intelligence. These datasets not only facilitate breakthroughs by enabling the training and evaluation of advanced models, but also provide standardized benchmarks for measuring and recognizing technological advancements. However, currently available smoke segmentation datasets \cite{yuan2019deep,yan2022transmission,yao2024fosp} face several limitations that restrict their effectiveness. First, datasets \cite{yuan2019deep,yan2022transmission} include a high proportion of synthetic images, introducing domain gaps between artificial and real-world scenarios, which may impair model performance in practical applications. Second, most datasets \cite{yan2022transmission,yao2024fosp} are relatively small in scale, with the largest publicly available dataset containing only about 6K real images. This limited size presents a significant challenge for training smoke segmentation models. Finally, many current datasets exhibit geographic bias. For instance, SmokeSeg~\cite{yao2024fosp} contains images exclusively from the United States. This narrow geographic scope can limit the generalization ability of models to diverse real-world environments. 

Our goal is to encourage the research community to push the frontiers of AI, promoting innovation in wildfire prevention and risk management, and ultimately contributing to environmental preservation and public safety. Australia is one of the countries most severely affected by wildfires \cite{andela2019global,tran2020high}. However, there is still no dedicated smoke segmentation dataset specifically focused on Australian wildfires. To address this gap, we collected real-world images from cameras mounted on Australian fire towers and handheld cameras. These images capture a wide range of environmental conditions, including both wildfires and planned burns. By using planned burns, we can reliably observe the very first traces of smoke under safe conditions, highlighting the critical need for early detection. Using these images, we constructed a wildfire smoke segmentation dataset, called AusSmoke, which contains more than 15K real images with fully labeled segmentation annotations.

\begin{table}[t]\centering
  \begin{tabular}{lccc}
    \toprule
    Name & Type & Size & Geography \\
    \midrule
    FESB-MLID \cite{bugaric2014adaptive} & Real & 400 & Single \\
    SYN70K \cite{yuan2019deep} & Synthetic & 73632 & - \\
    Smoke5K \cite{yan2022transmission} & Mixed & 5400 & Single \\
    SmokeSeg \cite{yao2024fosp} & Real & 6144 & Single \\
    \textbf{AusSmoke} & Real & 15248 & Single \\
    \textbf{MultiNatSmoke} & Real & 70818 & Multiple \\
    \bottomrule
  \end{tabular}
  \caption{Comparison of publicly available wildfire smoke segmentation datasets. FESB-MLID, Smoke5K, and SmokeSeg are limited in both scale and geographic diversity.  In contrast, our AusSmoke dataset provides over 15K real images collected exclusively from Australia. Building further, our MultiNatSmoke benchmark consolidates international public sources with newly curated smoke annotations, yielding more than 70K real images spanning diverse global environments.}
  \label{tab.dataset_comparison}
\end{table}

\begin{figure}[t]
  \centering
  \includegraphics[width=0.8\linewidth]{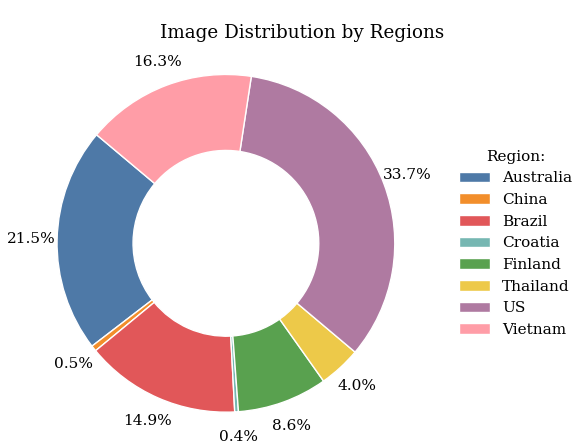}
  \caption{Region-wise distribution of the MultiNatSmoke. This wide geographic coverage demonstrates the global diversity of MultiNatSmoke, reducing regional bias and improving the dataset’s ability to support the development of models that generalize across varied environments.}
  \label{fig:geo}
\end{figure}

To address the limited geographic diversity of existing smoke segmentation datasets, we further introduce a large and diverse wildfire smoke segmentation benchmark, called MultiNatSmoke, by consolidating international public sources with newly provided smoke segmentation annotations. The benchmark contains 70,818 real smoke images, making it an order of magnitude larger than the previously largest publicly available real image smoke segmentation dataset. Table~\ref{tab.dataset_comparison} presents a statistical comparison between our benchmark and existing wildfire smoke segmentation datasets. Figure~\ref{fig:geo} illustrates the geographic distribution of our benchmark, highlighting its global diversity. 

In our benchmark, we evaluate a range of smoke segmentation models, including CNN-based approaches \cite{ronneberger2015u,chen2018encoder}, Transformer-based methods \cite{xie2021segformer,cheng2022masked,jain2023oneformer}, and smoke-specific architectures \cite{yan2022transmission,yao2024fosp}, to demonstrate their effectiveness and utility in advancing smoke segmentation. Experimental results show substantial improvements when existing models are trained on our dataset compared to the previously largest dataset. Furthermore, we assess the smoke segmentation models under zero-shot evaluation; that is, we train a model on certain datasets and then test its performance on other datasets that were not used during training. We observed that, with the same training effort, training data with geographical and contextual diversity achieve better results than training data derived from a single dataset. 

In summary, our contributions are threefold:
\begin{itemize}
    \item We present AusSmoke, a wildfire smoke segmentation dataset comprising over 15,000 real images collected across Australia. Because early fire detection is critical, the dataset emphasizes small-scale fires, with a significant portion sourced from planned burns.
    \item We introduce MultiNatSmoke, a large-scale wildfire smoke segmentation benchmark of real images, designed to overcome limitations in both dataset size and geographic diversity. The benchmark provides fully labeled pixel-level smoke segmentation annotations. 
    \item We benchmark our new dataset across CNN-based, Transformer-based, and smoke-specific models, showing substantial performance gains and improved generalization for early smoke detection.  
\end{itemize}

Together, these contributions advance the field of AI-driven wildfire response by providing the data infrastructure and benchmarking tools necessary to develop, evaluate, and deploy more accurate and generalizable early smoke detection systems in real-world settings.
\section{Related Work}
\paragraph{Wildfire Smoke Datasets.} 
Over the past two decades, the growing interest in wildfire monitoring has driven the development of numerous datasets including FIgLib \cite{dewangan2022figlib}, BoWFire \cite{chino2015bowfire}, Corsican \cite{toulouse2017computer}, FLAME1 \cite{shamsoshoara2021aerial}, and BA-UAV \cite{ribeiro2023burned}. These datasets support tasks ranging from smoke classification to smoke segmentation. To gain a comprehensive understanding of wildfire smoke-related datasets, we recommend consulting the review \cite{boroujeni2025fire}.  In this work, however, we focus primarily on datasets specifically designed for wildfire smoke segmentation \ie,~delineating smoke pixels in an image, which provides spatially explicit information crucial for early fire detection and automated scene understanding.  For smoke segmentation, SYN70K \cite{yuan2019deep} is a large synthetic set consisting of over 70,000 images, developed using advanced rendering techniques in Blender. It simulates diverse smoke scenarios across various environmental conditions. Although synthetic images can effectively represent smoke in controlled settings, they often fail to capture the full complexity of real-world smoke phenomena. To enhance real-world smoke segmentation, Smoke5K \cite{yan2022transmission} was introduced as a mixed dataset containing 1,000 real images and 4,000 synthetic images, with the synthetic subset sourced from SYN70K. \cite{yan2022transmission} demonstrated that models trained on Smoke5K achieved superior performance compared to those trained solely on SYN70K, highlighting the limitations of relying exclusively on synthetic data. However, the number of real images in Smoke5K remains relatively small. SmokeSeg \cite{yao2024fosp} comprises 6,144 real images, sourced primarily from FIgLib \cite{dewangan2022figlib}, and stands as the largest publicly available real-image smoke segmentation dataset with pixel-wise annotations. Despite its scale, all images in SmokeSeg originate from the United States, thereby limiting the dataset’s geographical diversity.

\vspace{1mm}
\noindent \textbf{Smoke Segmentation Models.} 
Recent object detection and segmentation \cite{ronneberger2015u,redmon2016you,he2017mask,li2018deep,chen2018encoder,xie2021segformer,zheng2022towards,cheng2022masked,ji2023deep,li2023rein,hong2024goss,chen2021channel,liu2022generalised,Shrestha2025SDIPaste,Qi2026SmokeBench} has attracted increasing attention, and smoke segmentation models \cite{yan2022transmission,yao2024fosp,Shrestha2025GIMO,Zhao2026FalseAlarm} have been proposed or adapted to address this challenging task. Among smoke-specific methods, Trans-BVM \cite{yan2022transmission} leverages a Bayesian generative framework to estimate posterior distributions of parameters and predictions while incorporating a transmission-guided local coherence loss to capture pairwise relationships between pixels. FoSp \cite{yao2024fosp}, on the other hand, represents a state-of-the-art approach on the SmokeSeg dataset by combining CNN-based feature extraction with attention mechanisms that enhance boundary precision, making it particularly effective for small-scale smoke structures. Beyond these dedicated models, general-purpose segmentation architectures have also been employed for smoke analysis. CNN-based methods such as U-Net \cite{ronneberger2015u} and DeepLabv3+ \cite{chen2018encoder} have demonstrated strong performance, with U-Net excelling on smaller datasets due to its skip-connected encoder-decoder design, while DeepLabv3+ exploits atrous convolutions and Atrous Spatial Pyramid Pooling for multi-scale feature representation. More recent transformer-based architectures have shown promising results on the SmokeSeg dataset \cite{yao2024fosp}, with SegFormer \cite{xie2021segformer} using a hierarchical ViT encoder and lightweight MLP decoder for efficient multi-scale feature extraction, and Mask2Former \cite{cheng2022masked} employing masked attention and learnable queries to unify semantic, instance, and panoptic segmentation tasks.

\section{Method}
\subsection{AusSmoke Dataset}

\begin{figure*}[t!]
  \centering
  \begin{minipage}[t]{0.48\textwidth}
    \centering
    \includegraphics[width=0.8\linewidth,height=5cm]{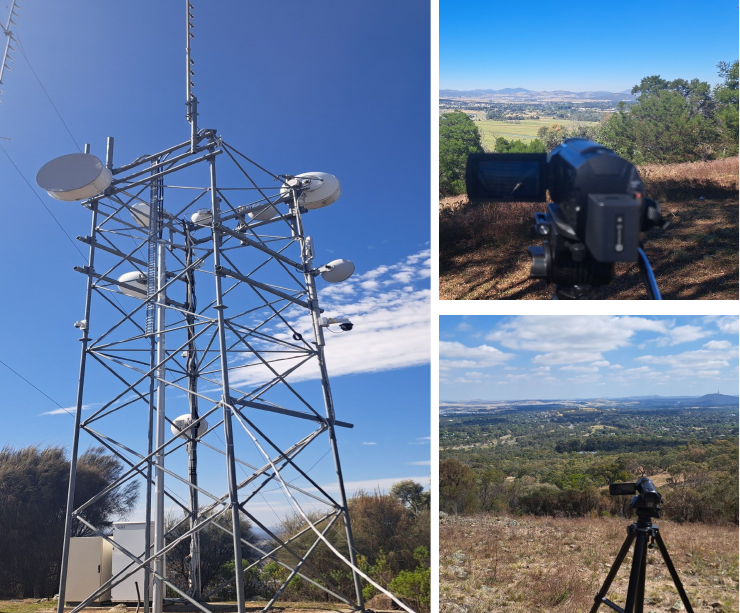}
    \caption{The cameras used for the AusSmoke dataset: on the left is the PTZ camera (Axis Q6318-LE); on the top right is the handheld Sony FDR-AX53 video camera; and on the bottom right is the Panasonic HC-VX1 video camera.}
    \label{fig:cameras}
  \end{minipage}\hfill
  \begin{minipage}[t]{0.48\textwidth}
    \centering
    \includegraphics[width=0.7\linewidth,height=5cm]{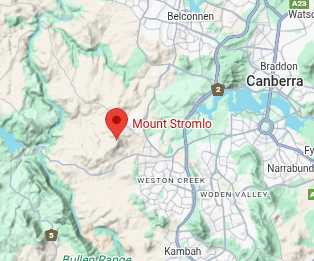}
    \caption{The PTZ camera used for data collection was located at Mount Stromlo (latitude -35.214, longitude 149.184) and operated on a continuous rotation to enable opportunistic smoke image capture.}
    \label{fig:stromlo}
  \end{minipage}
\end{figure*}

Wildfires pose a severe threat to Australia, making it one of the most affected regions \cite{andela2019global}. However, no smoke segmentation dataset has been developed for Australian wildfire scenarios. To address this gap, we present AusSmoke, a new wildfire smoke segmentation dataset constructed from real-world smoke images collected from Australia.

We collected new data from both unplanned and planned burns in the Australian Capital Territory (ACT), Australia, to obtain high-quality images of smoke during the early stages of fire development. These planned burns are managed by ACT Parks and Conservation Service and ACT Rural Fire Service. Through these burns, we can reliably capture the earliest visible indications of smoke. Smoke imagery was acquired using a combination of handheld and pan-tilt-zoom (PTZ) cameras. The PTZ camera (Axis Q6318-LE) (see Figure~\ref{fig:cameras}), permanently installed on a communications tower on Mount Stromlo (Latitude -35.214, Longitude 149.184, see Figure \ref{fig:stromlo}) was set on a permanent rotation for incidental smoke image collection. For some planned fires, the PTZ camera was oriented toward the burn prior to ignition. To complement this, handheld video cameras (Panasonic HC-VX1, and Sony fdr-ax53) (see Figure~\ref{fig:cameras}) were deployed at distances between approximately 1 and 30 kilometers from planned burns, prior to ignition. The AusSmoke database is unique in that it i) contains exclusively Australian images and ii) emphasises the collection of smoke images from small, establishing fires. In addition, the dataset also contains images provided by members of the Australian fire management community and from Australian-based fire detection cameras. Figure \ref{fig:teaser} illustrates the newly established AusSmoke dataset. In processing the AusSmoke data, we retained only the frames recorded during active burning periods, while discarding those irrelevant to the burning activity to ensure quality and relevance. Overall, we collected 15,248 images to construct the AusSmoke dataset.
\subsection{MultiNatSmoke Benchmark}

The performance of smoke detection approaches is constrained by the small scale of existing wildfire smoke datasets, the shortage of real images, and limited geographic diversity, which make the construction of a large-scale, international, real-world dataset essential. To address these challenges, we build a new smoke segmentation benchmark by combining our AusSmoke dataset with publicly available sources. We focus primarily on outdoor wildfire smoke scenes, excluding indoor and urban environments. The public datasets used in our benchmark are listed below. For the benchmark, we provide fully labeled pixel-level smoke segmentation masks. 

\noindent$\bullet$~\textbf{United States.} 
The FIgLib dataset \cite{dewangan2022figlib} consists of images captured by fixed-view cameras from the High Performance Wireless Research and Education Network (HPWREN) in Southern California, United States. The dataset was collected using 101 cameras deployed across 30 stations. Several datasets, including AI-for-Mankind \cite{aiformankind2020wildfire}, SmokeSeg \cite{yao2024fosp}, Smoke5K \cite{yan2022transmission}, and firecam \cite{govil2020preliminary}, offer annotations built upon the FIgLib data.

\noindent$\bullet$~\textbf{Finland.} 
Boreal-Forest-Fire \cite{raita2023combining,Pesonen2025Boreal} was collected during four controlled forest restoration burns conducted in the summer of 2022 in Finland. These burns were carried out in four Finnish towns: Evo (E25.1856, N61.2281), Heinola (E26.4425, N61.3008), Karkkila (E23.9781, N60.6422), and Ruokolahti (E28.9222, N61.3506). The footage comprises both close-range and long-range video captures, recorded using unmanned aerial vehicles (drones) equipped with Phantom P4 action cameras.

\noindent$\bullet$~\textbf{Thailand.} 
FireSpot \cite{pornpholkullapat2023firespot}  was collected through a collaboration between the National Electronics and Computer Technology Center (NECTEC) and three local municipalities, Pa Miang, Nong Yaeng, and Choeng Doi, in Chiang Mai, Thailand. Among the images, 2,817 contain smoke, with smoke regions precisely annotated using bounding boxes to facilitate detection tasks.

\noindent$\bullet$~\textbf{Brazil.} 
D-Fire \cite{de2022automatic} is an image dataset specifically designed for fire and smoke detection. It comprises a total of 5,867 smoke images with 11,865 annotated bounding boxes indicating smoke regions. The images were collected from multiple sources, including legal fire simulations conducted at the Technological Park of Belo Horizonte, Brazil, as well as from surveillance cameras monitoring landscapes at the Universidade Federal de Minas Gerais (UFMG) and the Serra Verde State Park, both located in Belo Horizonte.

\noindent$\bullet$~\textbf{Croatia.} 
FESB-MLID \cite{bugaric2014adaptive,braovic2017cogent} comprises 400 images of natural Mediterranean landscapes, accompanied by hand-labeled segmentation maps. This dataset was developed by the Faculty of Electrical Engineering and Naval Architecture (FESB) under the Wildfire Research Center at the University of Split, Croatia.

\noindent$\bullet$~\textbf{China.} 
The Forest Fire \cite{zhang2018wildland} dataset is primarily designed for monitoring wildfire scenes. It is constructed from video footage captured by surveillance cameras installed in lookout towers and uncrewed aerial vehicles (UAVs). The dataset was collected by the University of Science and Technology of China (USTC), China.

\noindent$\bullet$~\textbf{Vietnam.} 
WSDataset \cite{vu2023ung} was collected by the Research Team at IEC PTIT, Vietnam. It comprises 11,539 smoke-positive frames, each annotated with bounding boxes. Data collection was conducted using over forty high-altitude, high-resolution cameras with 360-degree rotation capabilities, strategically deployed across forested areas in Lam Dong Province, Vietnam.

\subsection{Data Annotation}

Annotating smoke is challenging because of its irregular boundaries, transparency, and small size in early stages. These properties make smoke blend into the background, producing diffuse and ambiguous edges that are difficult to separate from the environment. Our annotation team comprises the authors themselves, all of whom are experienced researchers in computer vision and wildfire studies. To overcome the challenges associated with smoke annotation, we use a multi-stage annotation approach to enhance accuracy and efficiency. For smoke images with existing segmentation annotations, such as SmokeSeg \cite{yao2024fosp} and Smoke5K \cite{yan2022transmission}, we retained the provided masks as ground truth following thorough manual verification. For datasets annotated with bounding boxes, such as D-Fire \cite{de2022automatic}, we employed the Segment Anything Model (SAM) \cite{kirillov2023segment} using bounding boxes as prompts to generate initial segmentation masks. The collected masks are then used to train a smoke segmentation model \cite{yao2024fosp}, which is further applied to unannotated smoke images to produce pseudo-labels. Once all images were equipped with either ground truth or pseudo-labels, we adopted a self-training strategy \cite{xie2020self} to refine segmentation performance. At each stage, we manually verified the automatically generated labels, removing poor segmentations. After the initial round of self-training, segmentation outputs were further refined through manual review. Around 7K low-quality results were manually re-annotated using SAM with new interactive prompts. 

\subsection{Statistics}

After data collection and annotation, our smoke segmentation benchmark comprises 70,818 images, including 55,570 images sourced from publicly available datasets and 15,248 images from our own AusSmoke dataset. We split the dataset into a training set of 59,735 images and a test set of 11,083 images. Table \ref{tab.dataset_stat} compares available wildfire smoke segmentation datasets based on data size. Smoke5K \cite{yan2022transmission} is a mixed dataset containing 5,400 images, including both real and synthetic samples. SmokeSeg \cite{yao2024fosp} comprises 6,144 real-world images with pixel-wise annotations. However, all of the real-image data for both datasets originates exclusively from the United States. In contrast, our newly collected real-world imagery from Australia includes 15,248 images and our benchmark introduces a substantial advancement, featuring 70,818 pixel-wise labeled real images collected from multiple international locations, significantly improving both dataset size and geographic diversity. 

\begin{table}[t]
  \centering
  \small 
  \setlength{\tabcolsep}{5.5pt} 
  \begin{tabular}{lrrrr}
    \toprule
    Name & Size & Small & Medium & Large \\
    \midrule
    Smoke5K \cite{yan2022transmission} & 5,400  & 15.9\% &  7.0\% & 77.1\% \\ 
    SmokeSeg \cite{yao2024fosp}        & 6,144  & 60.5\% & 27.6\% & 11.9\% \\ 
    \textbf{AusSmoke}                  & 15,248 & 63.8\% & 26.4\% &  9.8\% \\ 
    \textbf{MultiNatSmoke}             & 70,818 & 55.9\% & 18.9\% & 25.2\% \\ 
    \bottomrule
  \end{tabular}
  \caption{Comparison of wildfire smoke segmentation datasets in terms of total dataset size and distribution of smoke regions across three scales (Small, Medium, and Large). While Smoke5K is dominated by large-scale smoke regions, SmokeSeg, AusSmoke, and MultiNatSmoke contain mostly small regions. MultiNatSmoke offers a more balanced distribution between medium and large.}
  \label{tab.dataset_stat}
\end{table}

We analyze the smoke pixel ratio in images for smoke segmentation datasets that contain real images, following \cite{yao2024fosp}. The categories are defined based on the smoke pixel ratio $\delta$ in an image as follows: a region is classified as \textbf{Small} if $\delta < 0.5\%$, \textbf{Medium} if $0.5\% < \delta < 2.5\%$, and \textbf{Large} if $\delta > 2.5\%$. Table \ref{tab.dataset_stat} shows that Smoke5K \cite{yan2022transmission} exhibits a strong bias toward large smoke regions, with 77\% of its images fall into the Large category. In contrast, SmokeSeg \cite{yao2024fosp} focuses on smaller smoke regions to support early smoke segmentation with 60.55\% of its images classified as Small. Our Australian dataset is particularly challenging, with over 63\% of the images containing small-sized smoke regions. Our MultiNatSmoke presents a more balanced distribution with 55.91\% Small, 18.94\% Medium, and 25.15\% Large smoke pixels.
According to the source of the data, AusSmoke (15,248), FigLib (12,435), D-Fire (10,525), and kaggle-wildfire-smoke-detection (11,539) are the four largest subsets in terms of image quantity, and they collectively contribute over 54,000 images, accounting for more than 76\% of the total. 
\section{Experiments}
\label{sec:experiments}

\subsection{Segmentation Models}
To comprehensively evaluate the effectiveness of our smoke segmentation dataset, we conducted experiments using several representative segmentation models, including CNN-based architectures (U-Net \cite{ronneberger2015u} and DeepLabv3+ \cite{chen2018encoder}), transformer-based architectures (OneFormer \cite{jain2023oneformer}, SegFormer \cite{xie2021segformer} and Mask2Former \cite{cheng2022masked}), and smoke-specific models (Trans-BVM \cite{yan2022transmission} and FoSp \cite{yao2024fosp}). Notably, FoSp is reported to be a state-of-the-art method on SmokeSeg \cite{yao2024fosp}.

\subsection{Implementation Details}

We trained all models on two NVIDIA RTX 4090 GPUs, following the settings provided by open-source resources. All images were resized to 512 $\times$ 512. All models were trained for 20 epochs using AdamW \cite{adamw}, where weight decay was set to 0.01 and a learning rate of $1 \times 10^{-4}$, with early stopping. The batch size was set to 32. U-Net and DeepLabv3+ employed ResNet-50 backbones pretrained on ImageNet. SegFormer used an MiT-b2 backbone \cite{xie2021segformer} pretrained on ImageNet \cite{deng2009imagenet}, while Mask2Former used a Swin Transformer backbone \cite{liu2021swin} pretrained on ADE20K \cite{zhou2017scene}. All parameters were set to trainable during the training process, and we optimized the above models with binary cross-entropy loss.  We implemented all models in PyTorch (v2.6.0+cu124) with Python 3.9. U-Net and DeepLabv3+ were built using the \textit{segmentation models PyTorch} library \cite{Iakubovskii_2019}, while SegFormer and Mask2Former were instantiated via \textit{transformers} \cite{wolf-etal-2020-transformers}. FoSp and Trans-BVM were implemented using the official implementations provided by the authors (\url{https://github.com/LujianYao/FoSp} and \url{https://github.com/SiyuanYan1/Transmission-BVM}, respectively).

\subsection{Evaluation Metrics }
To quantitatively assess the accuracy and quality of the segmentation outputs, we used a suite of common evaluation metrics, including Intersection over Union (IoU), Mean Squared Error (MSE), F-measure, Precision and Recall. 

\textbf{Intersection over Union (IoU)} measures the overlap between the predicted smoke segmentation region and the ground truth region, and is defined as \( \text{IoU} = \frac{|A \cap B|}{|A \cup B|} \), where \( A \) is the predicted region and \( B \) is the ground truth region.

\textbf{Mean Square Error (MSE)} measures the average squared difference between the predicted and ground truth smoke segmentation masks, defined as \( \text{MSE} = \frac{1}{N} \sum_{i=1}^{N} (P_i - G_i)^2 \), where \( N \) is the total number of pixels, and \( P_i \) and \( G_i \) are the predicted and ground truth values at pixel \( i \), respectively. A lower MSE indicates higher segmentation accuracy.

\textbf{Precision} measures the proportion of correctly predicted positive pixels among all pixels classified as positive, defined as \( \text{Precision} = \frac{TP}{TP + FP} \), where \( TP \) and \( FP \) are the numbers of true and false positive pixels, respectively. High precision indicates that most predicted positives are correct.

\textbf{Recall} measures the proportion of correctly predicted positive pixels among all actual positives, given by \( \text{Recall} = \frac{TP}{TP + FN} \), where \( FN \) is the number of false negatives. High recall indicates that most actual positives are detected.

\textbf{F-1} is the harmonic mean of precision and recall, defined as \( F_\beta = (1 + \beta^2) \cdot \frac{\text{Precision} \cdot \text{Recall}}{(\beta^2 \cdot \text{Precision}) + \text{Recall}} \), where \( \beta \) adjusts the relative weight: \( \beta = 1 \) gives equal weight, \( \beta > 1 \) emphasizes recall, and \( \beta < 1 \) emphasizes precision. It is especially useful for imbalanced datasets.
\section{Results}
\label{sec:results}

\subsection{Main Results}

\begin{table}[t]
\centering
\resizebox{\linewidth}{!}{
\begin{tabular}{l |ccccc}
\toprule
\cmidrule(lr){2-6}
Method & $IoU\uparrow$ & $F_1\uparrow$ & $MSE\downarrow$ & $Prec\uparrow$ & $Rec\uparrow$ \\
\midrule
U-Net \cite{ronneberger2015u}  (MICCAI 2015)                & 70.50 & 82.17 & 0.0146          & 84.56 & 80.93 \\
DeepLabv3+ \cite{chen2018encoder}  (ECCV 2018)                 & 70.68 & 82.26 & 0.0132          & 82.03 & 83.60 \\
\arrayrulecolor{black!30}\midrule
Mask2Former \cite{cheng2022masked} (CVPR~2022)      & 70.92 & 82.21 & 0.0128          & 82.79 & 83.11 \\
SegFormer \cite{xie2021segformer} (NeurIPS~2021)     & \textbf{73.47} & \textbf{84.21} & \textbf{0.0118} & \textbf{85.18} & 84.20 \\
OneFormer \cite{jain2023oneformer} (CVPR~2023) & 69.46 & 80.90 & 0.0127 & 81.70 & 82.27 \\ 
\arrayrulecolor{black!30}\midrule
Trans-BVM \cite{yan2022transmission} (AAAI~2022) & 67.35 & 79.78 & 0.0155          & 82.65 & 79.78 \\
FoSp \cite{yao2024fosp} (AAAI~2024)             & 72.20 & 83.32 & 0.0126          & 82.85 & \textbf{84.92} \\
\arrayrulecolor{black}\bottomrule
\end{tabular}
}

\caption{Performance of segmentation models on the MultiNatSmoke test set. 
SegFormer achieves the best overall results across $IoU$, F\textsubscript{1}, MSE, and precision, while FoSp attains the highest recall. Notably, FoSp is reported to be a state-of-the-art method on SmokeSeg \cite{yao2024fosp}.}
\label{tab:main_results}
\end{table}

Table~\ref{tab:main_results} summarizes the performance of segmentation models on the MultiNatSmoke test set. SegFormer consistently outperforms the other models, posting the best $IoU$ (73.47), best $F_1$ (84.21), lowest MSE (0.0118), and the highest precision (85.18); its recall (84.20) is second only to FoSp, which achieves the top recall at 84.92 while remaining competitive on $IoU$/F\textsubscript{1}/MSE (72.20/83.32/0.0126). DeepLabv3+ and Mask2Former form a strong middle tier with similar $IoU$ ($\approx$70.7–70.9), $F_1$ ($\approx$82.2–83.1), and MSE (0.0132–0.0128). U-Net exhibits a precision–recall tradeoff, high precision (84.56) but lower recall (80.93), yielding $IoU$ 70.50. Trans-BVM trails the group across metrics ($IoU$ 67.35, $F_1$ 79.78, MSE 0.0155).

\subsection{Impact of Smoke Size on Model Performance}

\begin{table*}[!ht]
\centering
\setlength{\tabcolsep}{4pt}
\resizebox{\linewidth}{!}{%
\begin{tabular}{l ccccc ccccc ccccc}
\toprule
\multirow{2}{*}{Method}
& \multicolumn{5}{c}{Small}
& \multicolumn{5}{c}{Medium}
& \multicolumn{5}{c}{Large} \\
\cmidrule(lr){2-6}\cmidrule(lr){7-11}\cmidrule(lr){12-16}
& $IoU\uparrow$ & $F_1\uparrow$ & $MSE\downarrow$ & $Prec\uparrow$ & $Rec\uparrow$
& $IoU\uparrow$ & $F_1\uparrow$ & $MSE\downarrow$ & $Prec\uparrow$ & $Rec\uparrow$
& $IoU\uparrow$ & $F_1\uparrow$ & $MSE\downarrow$ & $Prec\uparrow$ & $Rec\uparrow$ \\
\midrule
U\text{-}Net       & 59.99 & 74.34 & \textbf{0.0007} & \textbf{74.62} & 75.57 & 60.70 & 74.50 & 0.0052 & 76.50 & 74.78 & 75.65 & 85.77 & 0.0413 & 88.35 & 83.58 \\
DeepLabv3+         & 57.71 & 72.48 & 0.0008          & 70.18 & 76.68 & 59.43 & 73.62 & 0.0055 & 72.68 & 77.06 & 77.64 & 87.16 & 0.0373 & 87.46 & 87.04 \\
\arrayrulecolor{black!30}\midrule
Mask2Former        & 58.61 & 72.95 & 0.0008          & 73.41 & 75.16 & 61.22 & 74.71 & 0.0055 & 78.19 & \textbf{87.19} & 77.89 & 87.19 & 0.0360 & 87.58 & 87.26 \\
SegFormer          & \textbf{61.73} & \textbf{75.78} & \textbf{0.0007} & 74.44 & 78.59 & \textbf{65.30} & \textbf{78.25} & \textbf{0.0045} & \textbf{78.73} & 79.66 & \textbf{79.57} & \textbf{88.37} & \textbf{0.0336} & \textbf{89.45} & \textbf{87.49} \\
OneFormer & 58.14 & 72.40 & 0.0009 & 70.78 & 77.08 & 61.96 & 75.22 & 0.0055 & 76.52 & 77.47 & 77.87 & 87.06 & 0.0357 & 87.57 & 87.31 \\
\arrayrulecolor{black!30}\midrule
Trans\text{-}BVM   & 52.53 & 68.35 & 0.0010          & 63.90 & 75.50 & 56.61 & 71.39 & 0.0061 & 69.48 & 76.52 & 75.33 & 85.52 & 0.0440 & 85.37 & 86.09 \\
FoSp               & 58.95 & 73.40 & 0.0008          & 68.80 & \textbf{80.62} & 63.52 & 76.93 & 0.0048 & 73.40 & 82.38 & 78.75 & 87.81 & 0.0353 & 89.16 & 86.71 \\
\arrayrulecolor{black}\bottomrule
\end{tabular}%
}
\caption{Performance of segmentation models on the MultiNatSmoke test set, evaluated by smoke size (Small, Medium, Large). Across all methods, larger smokes are generally easier to segment, with $IoU$ and $F_1$ improving from Small to Large. SegFormer delivers the most consistent and balanced results, achieving the best overall performance across nearly all metrics and sizes. FoSp stands out for its strong recall, particularly on Small and Medium cases.
}
\label{tab:results_size}
\end{table*}

We analyse how segmentation performance varies with the size of the smoke. Table~\ref{tab:results_size} presents comparative results of segmentation methods on the MultiNatSmoke test set, evaluating their performance across different object sizes: Small, Medium, and Large. The results show that larger smoke plumes are easier to segment across all models, with $IoU$ and $F_1$ rising from Small to Large while MSE grows (as expected with more foreground pixels). SegFormer is consistently strongest: it tops Small $IoU$/$F_1$ and matches U-Net for the best Small MSE (0.0007); on Medium it achieves the best $IoU$, $F_1$, MSE, and Precision, yielding the most balanced performance, though Mask2Former attains the highest Recall (87.19) with FoSp second (82.38); on Large SegFormer sweeps all five metrics ($IoU$ 79.57, $F_1$ 88.37, MSE 0.0336, Precision 89.45, Recall 87.49). FoSp stands out for recall-oriented behavior—best Recall on Small (80.62), second-best MSE on Medium/Large, and competitive Precision (notably 89.16 on Large). U-Net tends toward higher Precision (best on Small, strong on Large) but lags in Recall, reflecting a conservative segmentation bias. DeepLabv3+ and Mask2Former form a solid middle tier, with Mask2Former particularly effective at recovering positives on Medium/Large. Trans-BVM trails on most metrics and sizes. Overall, transformer-based decoders (especially SegFormer) deliver the best accuracy–error trade-off, while FoSp maximizes sensitivity.

\subsection{Impact of Data Scale on Model Performance}

\begin{table*}[t]
\centering
\resizebox{\linewidth}{!}{%
\begin{tabular}{l|ccc|ccc|ccc|ccc}
\toprule
\multirow{2.5}{*}{Dataset} 
& \multicolumn{3}{c|}{Small} 
& \multicolumn{3}{c|}{Medium} 
& \multicolumn{3}{c|}{Large} 
& \multicolumn{3}{c}{Total} \\
\cmidrule(l){2-4} \cmidrule(l){5-7} \cmidrule(l){8-10} \cmidrule(l){11-13}
& $IoU \uparrow$ & $F_1 \uparrow$ & $MSE \downarrow$
& $IoU \uparrow$ & $F_1 \uparrow$ & $MSE \downarrow$
& $IoU \uparrow$ & $F_1 \uparrow$ & $MSE \downarrow$
& $IoU \uparrow$ & $F_1 \uparrow$ & $MSE \downarrow$ \\
\midrule
Smoke5K  & 17.92 & 28.84 & 0.0047 & 32.83 & 47.55 & 0.0168 & 59.73 & 74.13 & 0.0900 & 43.84 & 58.33 & 0.0334 \\
SmokeSeg & 42.61 & 59.69 & 0.0017 & 62.50 & 76.88 & 0.0039 & 62.86 & 77.13 & 0.0246 & 61.16 & 75.63 & 0.0093 \\
\midrule
MultiNatSmoke-20\%     & 55.34 & 70.60 & 0.0009 & 61.38 & 75.06 & 0.0056 & 76.97 & 86.69 & 0.0399 & 69.83 & 81.65 & 0.0141 \\
MultiNatSmoke-40\%     & 60.70 & 74.84 & 0.0007 & 62.36 & 75.96 & 0.0049 & 78.23 & 87.51 & 0.0366 & 72.21 & 83.35 & 0.0128 \\
MultiNatSmoke-60\%     & 60.04 & 74.25 & 0.0008 & 63.15 & 76.62 & 0.0047 & 78.53 & 87.68 & 0.0363 & 72.54 & 83.57 & 0.0128 \\
MultiNatSmoke-80\%     & 60.58 & 74.88 & 0.0007 & 65.17 & 78.24 & \textbf{0.0043} & 78.72 & 87.81 & 0.0356 & 72.73 & 83.71 & 0.0124 \\
MultiNatSmoke-full    & \textbf{61.73} & \textbf{75.78} & \textbf{0.0007} & \textbf{65.30} & \textbf{78.25} & 0.0045 & \textbf{79.57} & \textbf{88.37} & \textbf{0.0336} & \textbf{73.47} & \textbf{84.21} & \textbf{0.0118} \\
\bottomrule
\end{tabular}
}
\caption{Performance of SegFormer on the MultiNatSmoke test set when trained with different datasets and varying proportions of MultiNatSmoke. 
Training on MultiNatSmoke consistently yields the best results across all smoke sizes, with performance steadily improving as more of the dataset is used, highlighting the effectiveness and scalability of the proposed dataset. }
\label{tab:segformer_performance_diff_datasets}
\end{table*}

Table \ref{tab:segformer_performance_diff_datasets} shows the impact of training data on the performance of SegFormer when evaluated on the MultiNatSmoke test set. Among the three training datasets, the model trained on MultiNatSmoke (60K real images) consistently outperforms those trained on Smoke5K (5K mixed images) and SmokeSeg (about 6K real images) across all smoke region sizes. Notably, for small smoke, MultiNatSmoke achieves the highest $IoU$ of 61.73 and $F_1$ of 75.78, while also producing the lowest MSE ($0.0007$), indicating superior localization and segmentation precision. This trend continues for medium and large regions, with especially strong performance on large regions ($IoU = 79.57$, $F_1 = 88.37$), suggesting improved robustness in detecting dense smoke. In contrast, the model trained on Smoke5K shows significantly lower performance, particularly on small regions, highlighting the limitations of real-world data with limited variability. SmokeSeg performs moderately well, but still lags behind our dataset, especially in large-scale smoke scenarios. We further examine the effect of training data scale by evaluating SegFormer trained on incremental subsets (20\%, 40\%, 60\%, 80\%) of the MultiNatSmoke dataset. The results show a steady improvement in performance as the training set size increases. 

\begin{figure*}[t!]
  \centering
  \includegraphics[width=0.9\linewidth,height=0.275\textheight]{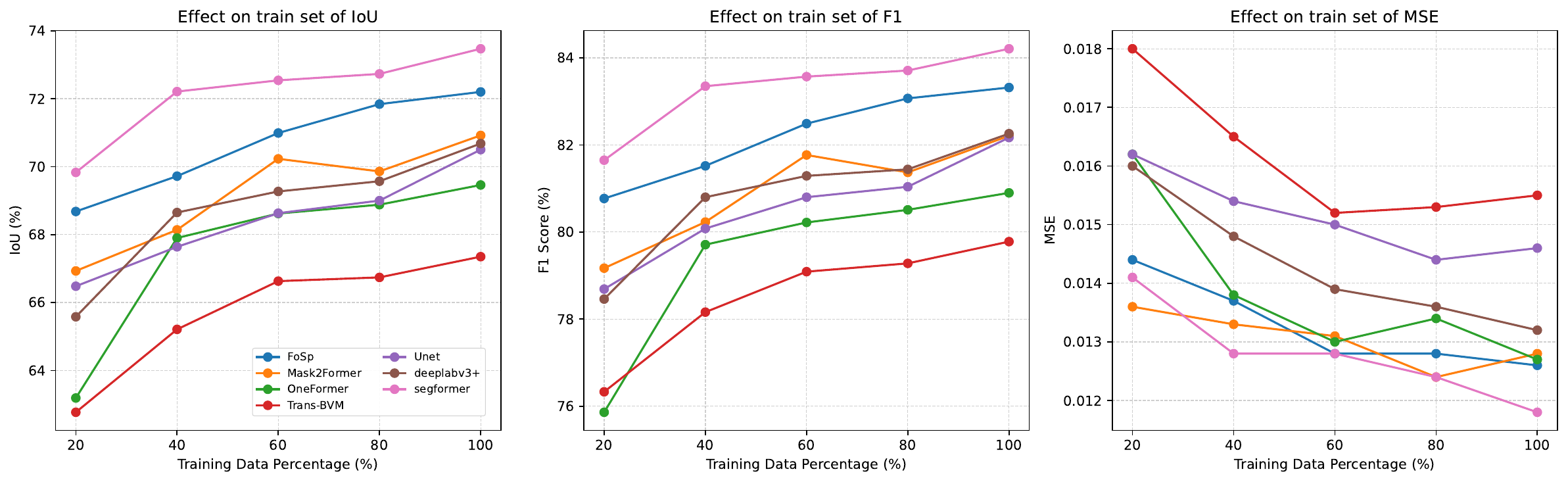}
  \caption{Comparison of smoke segmentation models at different training data scales. Larger training sets boost performance across models, with SegFormer leading consistently.}
  \label{fig:data_scale}
\end{figure*}

Figure \ref{fig:data_scale} demonstrates that, under different training data percentages, various smoke segmentation models reveal a consistent trend: as the proportion of training data increases, the smoke segmentation models generally achieve better performance across all metrics. Specifically, SegFormer consistently outperforms other models, achieving the highest $IoU$ and $F_1$ scores and the lowest MSE across all data scales. U-Net and DeepLabv3+ show steady improvements with increased data but lag behind SegFormer and Mask2Former. The FoSp model also shows promising results, outperforming the older CNN-based methods. Trans-BVM demonstrates the lowest performance overall, particularly at smaller data scales, suggesting it may be more sensitive to limited training data. Overall, the results underscore the importance of data availability in achieving high segmentation accuracy.

\subsection{Impact of Geo-Diversity on Model Performance}

\begin{table}[!ht] 
\centering 
\resizebox{\linewidth}{!}{%
\begin{tabular}{l|ccc|ccc} 
\toprule 
\multirow{2.5}{*}{Dataset} 
& \multicolumn{3}{c|}{Boreal} 
& \multicolumn{3}{c}{AusSmoke} \\ 
\cmidrule(l){2-4} 
\cmidrule(l){5-7} 
& $IoU\uparrow$ & $F_1\uparrow$ & $MSE\downarrow$ 
& $IoU\uparrow$ & $F_1\uparrow$ & $MSE\downarrow$ \\ 
\midrule 
SmokeSeg & 72.80 & 84.21 & 0.0712 & 39.34 & 56.02 & 0.0072 \\ 
MultiNatSmoke-sub & \textbf{78.78} & \textbf{88.07} & \textbf{0.0525} & \textbf{47.03} & \textbf{63.53} & \textbf{0.0057} \\ 
\bottomrule 
\end{tabular}%
} 
\caption{SegFormer performance on Boreal and AusSmoke when trained on SmokeSeg versus a size-matched subset of MultiNatSmoke. 
The geographically diverse MultiNatSmoke-sub achieves consistently higher $IoU$ and F\textsubscript{1} scores and lower MSE on both benchmarks.}
\label{tab:geographical_diversity} 
\end{table}

To evaluate the impact of geographical diversity in training data, we randomly selected a subset of 5K images from the MultiNatSmoke dataset, excluding the Boreal (UAV imagery) and AusSmoke (high proportion of small smoke), to match the size of the SmokeSeg dataset. We then trained SegFormer on both the SmokeSeg dataset and the MultiNatSmoke-sub, and evaluated performance on the Boreal and AusSmoke test sets. Table~\ref{tab:geographical_diversity} shows that on the Boreal dataset, MultiNatSmoke achieves significantly higher $IoU$ (78.78 vs.~72.80) and F\textsubscript{1} score (88.07 vs.~84.21), along with a notably lower MSE (0.0525 vs.~0.0712), indicating more accurate and consistent segmentation of smoke regions. This substantial margin highlights MultiNatSmoke-sub’s robustness in forested and possibly UAV-view smoke scenes. The performance gap is even more meaningful on the AusSmoke dataset, which presents a more challenging scenario due to the presence of small and faint smoke regions. MultiNatSmoke-sub significantly surpasses SmokeSeg in this context, with an $IoU$ of 47.03 and $F_1$ score of 63.53, compared to 39.34 and 56.02 for SmokeSeg. Moreover, it achieves a lower MSE (0.0057 vs.~0.0072), indicating greater precision in segmenting small smoke. These results underscore the effectiveness of MultiNatSmoke’s geographical diversity in enabling robust segmentation performance across varied regional smoke conditions.
\section{Conclusion}

We tackle key shortcomings of existing wildfire smoke segmentation datasets by introducing newly collected imagery from Australia alongside a substantially larger and more diverse benchmark. Our benchmark integrates international public sources with Australian data, increasing the scale of available datasets by a factor of ten. Through extensive evaluation across multiple smoke segmentation models, we show that models trained on our dataset achieve stronger performance and generalization, particularly in geographically diverse scenarios. These findings underscore the critical role of dataset scale and diversity in advancing AI-driven smoke detection systems. At the same time, the results reveal new research challenges: even the best-performing models achieve an IoU below 74\%, with performance dropping below 62\% for small smoke regions.

\paragraph{Acknowledgement} This research was in part supported by the ANU-Optus Bushfire Research Centre of Excellence. We thank Ian Tanner for the contributions to data collection.
{
    \bibliographystyle{ieeenat_fullname}
    \bibliography{main}
}

\end{document}


\maketitle

\section{Dataset Details}

We collected multiple public smoke datasets and our AusSmoke dataset to construct a diverse smoke segmentation benchmark. Table \ref{tab:data_stat_appendix} reports the number of samples in each data source, where the meaning of each column is as follows:

\begin{itemize}
    \item Train: images used for training,
    \item Test Small/Medium/Large: test subsets stratified by scale,
    \item Test Total: sum of all three test subsets,
    \item Total: overall image count (training + all test subsets).
\end{itemize}

\begin{table*}[ht]
\centering
\caption{Dataset statistics by source}
\label{tab:data_stat_appendix}
\scalebox{0.88}{%
  \begin{tabular}{lrrrrr|r}
    \toprule
    \textbf{DataSource}                         & \textbf{Train} & \textbf{Test Small} & \textbf{Test Medium} & \textbf{Test Large} & \textbf{Test Total} & \textbf{Total} \\
    \midrule
    FireSpot                                    &  2436 &  144 &   89 &  196 &   429 &  2865 \\
    Kaggle Forest Fire                          &   323 &    3 &   12 &   51 &    66 &   389 \\
    D-Fire                                      &  8444 &  395 &  333 & 1353 &  2081 & 10525 \\
    SmokeSeg                                    &  5644 &  181 &  168 &  151 &   500 &  6144 \\
    Smoke5K\_real                               &  1058 &  242 &   99 &   59 &   400 &  1458 \\
    firecam                                     &  1412 &  247 &    2 &    0 &   249 &  1661 \\
    FIgLib                                      & 10570 & 1420 &  336 &  109 &  1865 & 12435 \\
    Boreal                                      &  4905 &    0 &   14 & 1191 &  1205 &  6110 \\
    AI-for-Mankind                              &  1797 &  333 &   51 &   10 &   394 &  2191 \\
    FESB-MLID                                   &   203 &    0 &    0 &   50 &    50 &   253 \\
    kaggle-wildfire-smoke-detection             & 10439 & 1014 &   86 &    0 &  1100 & 11539 \\
    AusSmoke                                     & 12504 & 1740 &  722 &  282 &  2744 & 15248 \\
    \midrule
    \textbf{Total}                              & 59735 & 5719 & 1912 & 3452 & 11083 & 70818 \\
    \bottomrule
  \end{tabular}
}
\end{table*}

\vspace{1mm}
\noindent
Table \ref{tab:data_stat_appendix} and Table \ref{tab:image_counts_app} reveals that the MultiNatSmoke dataset we have proposed is diverse enough to effectively evaluate the performance of the models and could contribute to future research. It contains a total of 70,818 images, 59,735 of which are for training, and the test set contains 11,083 images across eight countries. In the test set, there are 5,719 small-scale (Test Small) samples. According to the source of the data, AusSmoke (15,248), FigLib (12,435), D-Fire (10,525), and kaggle-wildfire-smoke-detection (11,539) are the four largest subsets in terms of image quantity, and they collectively contribute over 54,000 images, accounting for more than 76\% of the total. Additionally, the distribution of the dataset across the source scale and test set is also significantly different. For example, Boreal is relatively concentrated in the large-scale test set (1,191 images), while firecam has almost no large-scale samples (0 images).

\begin{table*}[ht]
  \centering
  \caption{Number of Images per Country}
  \label{tab:image_counts_app}
  \begin{tabular}{l r r r r r r r r}
    \toprule
                  & Australia & China & Brazil & Croatia & Finland & Thailand & US    & Vietnam \\
    \midrule
    Image Count  &     15248 &   389 &  10525 &     251  &   6110  &    2865  & 23891 &   11539 \\
    \bottomrule
  \end{tabular}
\end{table*}

\section{Annotation Visualization}

\begin{figure*}[htb]
  \centering
  \includegraphics[width=0.98\linewidth]{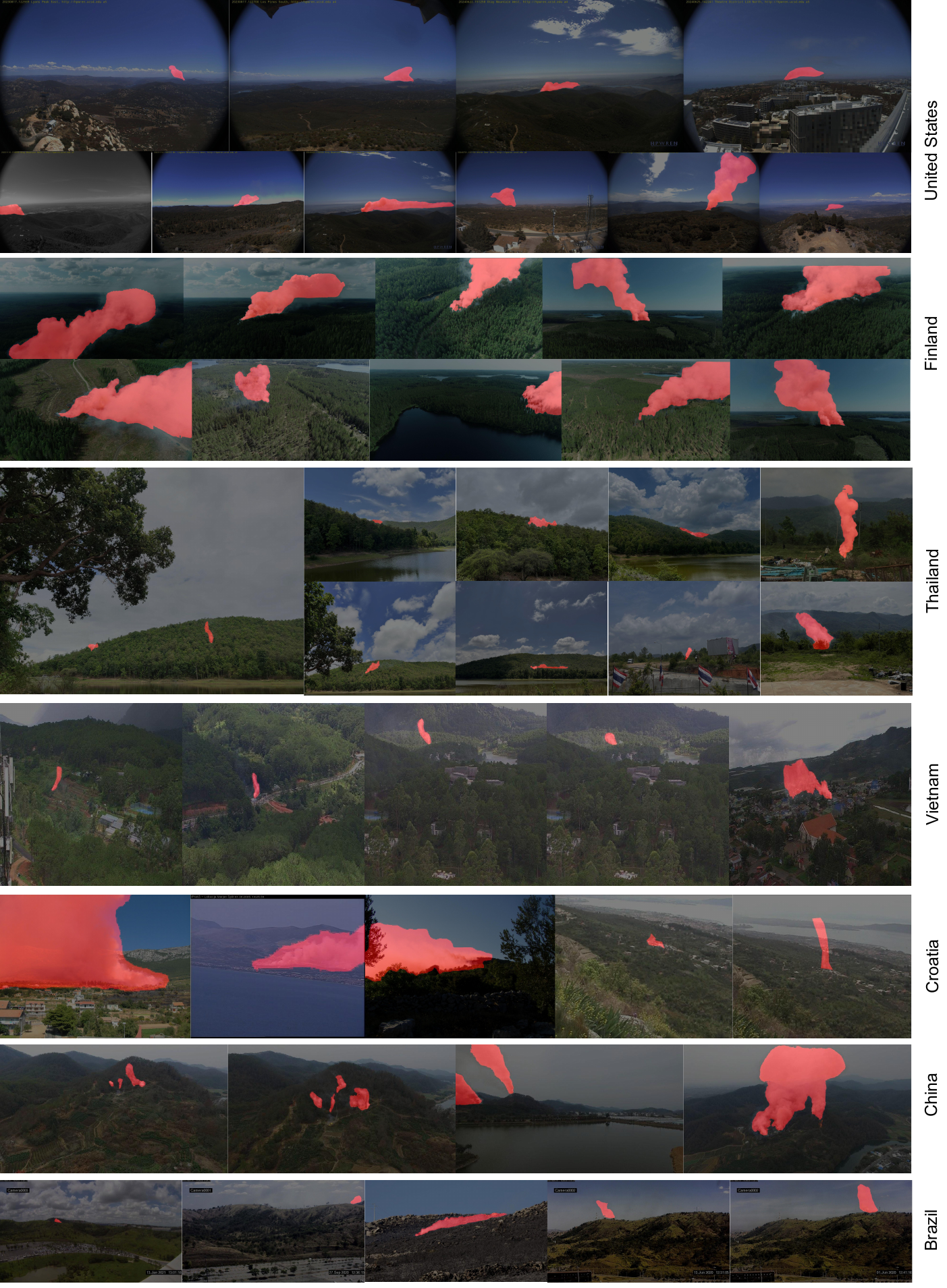}
  \caption{Example images with smoke segmentation annotation from our collected public data.}
  \label{fig.public_data}
\end{figure*}

In Figure~\ref{fig.public_data}, we show example smoke images together with their segmentation annotations, collected from public data for the MultiNatSmoke dataset.

